\documentclass[sigconf]{aamas}  %

\AtBeginDocument{%
  \providecommand\BibTeX{{%
    \normalfont B\kern-0.5em{\scshape i\kern-0.25em b}\kern-0.8em\TeX}}}

\usepackage{booktabs}    
\usepackage{flushend} %

\usepackage[utf8]{inputenc} %
\usepackage[T1]{fontenc}    %
\usepackage{hyperref}       %
\usepackage{url}            %
\usepackage{booktabs}       %
\usepackage{amsfonts}       %
\usepackage{nicefrac}       %
\usepackage{microtype}      %
\usepackage{dsfont}
\usepackage{graphicx}
\usepackage{subfigure}
\usepackage{amsmath}
\usepackage{bbm}
\usepackage{mathtools, nccmath}
\usepackage{wrapfig}
\usepackage{xcolor}
\setcopyright{ifaamas}  %
\copyrightyear{2020} %
\acmYear{2020} %
\acmDOI{} %
\acmPrice{} %
\acmISBN{} %
\acmConference[AAMAS'20]{Proc.\@ of the 19th International Conference on Autonomous Agents and Multiagent Systems (AAMAS 2020)}{May 9--13, 2020}{Auckland, New Zealand}{B.~An, N.~Yorke-Smith, A.~El~Fallah~Seghrouchni, G.~Sukthankar (eds.)}  %
\newcommand\given[1][]{\:#1\vert\:}

\begin{document}

\title{Likelihood Quantile Networks for Coordinating Multi-Agent Reinforcement Learning}  %
\author{Xueguang Lyu}
\affiliation{%
  \institution{Northeastern University}
  \city{Boston} 
  \state{MA} 
  \postcode{02115}
}
\email{lu.xue@northeastern.edu}
\author{Christopher Amato}
\affiliation{%
  \institution{Northeastern University}
  \city{Boston} 
  \state{MA} 
  \postcode{02115}
}
\email{c.amato@northeastern.edu}

\begin{abstract}
When multiple agents learn in a decentralized manner, the environment appears non-stationary from the perspective of an individual agent due to the exploration and learning of the other agents. 
Recently proposed deep multi-agent reinforcement learning methods 
have tried to mitigate this non-stationarity 
by attempting to determine which samples are from other agent exploration or suboptimality and take them less into account during learning.
Based on the same philosophy, this paper introduces a decentralized quantile estimator, which aims to improve performance by distinguishing non-stationary samples based on the likelihood of returns.
In particular, 
each agent considers the likelihood that other agent exploration and policy changes are occurring, essentially utilizing the agent's own estimations to weigh the learning rate that should be applied towards the given samples.
We introduce a formal method of calculating differences of our return distribution representations and methods for utilizing it to guide updates.
We also explore the effect of risk-seeking strategies for adjusting learning over time and propose adaptive risk distortion functions which guides risk sensitivity.
Our experiments, on traditional benchmarks and new domains, show our methods are more stable, sample efficient and more likely to converge to a joint optimal policy than previous methods.
\end{abstract}

\maketitle

\section{Introduction}

Many multi-agent reinforcement learning (MARL) methods have been developed (e.g., recent deep methods \cite{Foerster16,Foerster17,omidshafiei2017deep,Foerster18,Lowe17,Rashid18,palmer2018lenient,AAAI19}), but many of these approaches assume centralized training and decentralized execution.
Unfortunately, many realistic multi-agent settings will consist of agents that must (continue to or completely) learn online. 
In this case, each agent will be an Independent Learner which learns and executes in a decentralized manner using only its own sensor and communication information.
This decentralization can be more scalable and is necessary in cases where online (decentralized) learning takes place.
Unfortunately, with high probability, the decentralized learning agents will not converge to an optimal joint policy, but only optimal independent policies under the effect of {\it environment non-stationarity} caused by other agents' optimal independent policies \cite{fulda2007predicting}. 
In other words, %
without considering other agent exploration, agents will not typically 
achieve high performance.

Previous work, based on hysteretic Q-Learning \cite{matignon2007hysteretic} and leniency \cite{panait2006lenient}, which limit negative value updates (possibly due to exploration) %
have shown success in Deep reinforcement learning \cite{omidshafiei2017deep,palmer2018lenient} based on Deep Q-Networks (DQN) \cite{mnih2015human}.
Both approaches inject optimism: limiting the decrease of value estimations to alleviate the effect of other agents' exploration strategies and to encourage exploration outside of equilibria which easily trap agents without any injected optimism.
The trade-off between environment stochasticity biases and the aforementioned value overestimation, is considered inevitable since domain stochasticity and teammate policy shifts are traditionally indistinguishable.
Empirically, leniency shows higher learning stability compared to hysteretic learning, primarily due to a temperature-enabled leniency at different stages of estimation maturity \cite{palmer2018lenient}.
The leniency decay allows for a more faithful representation of domain dynamics during later stages of training, where it is probable that teammate policies become stable and near-optimal, assuming the rate of decay is appropriate and value maturity is synchronized across all states. 
Nevertheless, both hysteresis and leniency show only limited performance improvement and leniency introduces hyper-parameters that are hard to tune for new environments.
A recent study proposes incorporating negative update intervals \cite{palmer2018negative} to ignore sub-par episodes in a gradually relaxed way. But this method is designed for two-player temporally extended team games and requires an oracle mapping trajectories to domain specific `meta-actions,' making it inapplicable to the general multi-agent learning setting. 

Our work aims to develop a general method for improving decentralized multi-agent reinforcement learning. 
The method not only automatically identifies transitions involving sub-optimal teammate policies, especially explorations, but also adaptively schedules the amount of optimism applied to each training sample based on estimated value maturity, achieving improved performance without hyper-parameter interventions~\cite{omidshafiei2017deep,palmer2018lenient},  scheduling tables~\cite{palmer2018lenient} or specialized experience buffers~\cite{palmer2018negative}. In particular, we develop a novel method that extends a state-of-the-art deep distributional single-agent RL method \cite{dabney2018implicit}, Implicit Quantile Networks (IQN), to multi-agent settings to improve training stability %
and show how the auxiliary value distribution expectations can be used to identify exploratory teammates through what we call Time Difference Likelihood (TDL). 
TDL, uses distribution information to identify individual sub-par teammate explorations and guides the amount of optimism injected into the Q distribution; 
we call the new architecture Likelihood IQN.
We show empirically that our method is more robust even in domains that are  difficult for previous methods.
In addition, we propose a Dynamic Risk Distortion operator, in which risk distortion techniques can be applied in a scheduled fashion to produce optimistic policies that are robust to environment non-stationarity.

\section{Background}

We start by providing a summary of the background literature. This section includes introductions to MDPs, decentralized POMDPs and DQN, as well as a brief discussion on the challenges of independent learning in decentralized POMDPs.

\subsection{MDPs and Deep Q-Networks}

A Markov Decision Process (MDP) is defined with tuple $\langle \mathcal{S}, \mathcal{A}, \mathcal{T}, \mathcal{R} \rangle$, where $\mathcal{S}$ is a state space, $\mathcal{A}$ an action space, $\mathcal{T}(s,a,s^{\prime})$ the probability of transitioning from state $s{\in}\mathcal{S}$ to $s^{\prime}{\in}\mathcal{S}$ by taking action $a{\in}\mathcal{A}$, and $\mathcal{R}(s,a,s^{\prime})$ is the immediate reward for such a transition. 
The problem is to find an optimal policy $\pi^\star: \mathcal{S} \rightarrow \mathcal{A}$ which maximizes the expected sum of rewards (i.e., values) over time. 

Deep Q-Networks \cite{mnih2015human} consider a common practice where a nonlinear function approximator is used for estimating values by parameterizing the $Q$ function $Q^{\theta}(s,a)$ with parameters $\theta$ using a deep neural network, where $Q(s,a)$ is the expected maximum sum of rewards achievable in the future given state $s$ and action $a$.

DQN uses experience replay \cite{lin1993reinforcement} where each transition is stored in a fixed-sized experience buffer 
\begin{equation*}
    D_t = \big\{(s_1, a_1, r_1, s_2),...,(s_t, a_t, r_t, s_{t+1})\big\}
\end{equation*}
from which all training batches for the network are uniformly sampled to balance the network's tendency to bias towards more recent samples. The update of the network follows the following loss function:
\begin{equation}
    L_i(\theta_i) = \underset{s,a,r,s^{\prime} \sim U(D)}{\mathbb{E}}[(r+\gamma \max\limits_{a^{\prime}}Q^{\theta_i^-}(s^{\prime},a^{\prime})-Q^{\theta_i}(s,a))^2]
\end{equation}
where $\theta_i^-$ is the parameters for target network, an identical network whose parameters are not updated, but copied from the main network every $C$ steps as to maintain value stability.

\subsection{Decentralized POMDPs (Dec-POMDPs)}
General cooperative multi-agent reinforcement learning problems can be represented as 
decentralized partially observable Markov decision processes (Dec-POMDPs)~\cite{Book16}.  %
In a Dec-POMDP, each agent has a set of actions and observations, but there is a joint reward function and agents must choose actions based solely on their local observations.
A Dec-POMDP is defined as: $\langle \mathcal{I},  \mathcal{S},\mathcal{\boldsymbol{A}^I},\mathcal{Z},\mathcal{T},\mathcal{\boldsymbol{O}^I},\mathcal{\boldsymbol{R}} \rangle$ where $\mathcal{I}$ is a finite set of agents, $\mathcal{\boldsymbol{A}}^{i}$ is the action space for agent $i \in \mathcal{I}$, and $\mathcal{\boldsymbol{O}}^{i}$ is observation space of agent $i$. 
At every time step, a joint action $\boldsymbol{a} = \langle a^1,...,a^{|\mathcal{I}|} \rangle$ is taken, each agent sees it's own observation $o^i$, and all agents receive joint rewards based on the joint action  $\mathcal{\boldsymbol{R}} (s,\boldsymbol{a})$.

Earlier work has extended deep RL methods to partially observable MDPs (POMDPs)~\cite{Kaelbling98} and Dec-POMDPs. 
For instance, Deep Recurrent Q-Networks (DRQN)~\cite{hausknecht2015deep}   extends DQN to partially observable (single-agent) tasks, where a recurrent layer (LSTM) \cite{hochreiter1997long} was used to replace the first post-convolutional fully-connected layer of DQN. 
Hausknecht and Stone argue that the recurrent layer is able to integrate an arbitrarily long history which can be used to infer the underlying state. %
DQN and DRQN form the basis of many deep MARL algorithms (e.g., \cite{Foerster16,omidshafiei2017deep,palmer2018lenient}). Our work's basis is IQN (which we discuss later), and the recurrent version that we call IRQN.

Multi-Agent Reinforcement Learning  methods are 
 usually classified into two classes: Independent Learners (ILs) and Joint Action Learners (JALs) \cite{claus1998dynamics}. 
ILs observe only local actions $a^i$ for agent $i$, whereas JALs have access to joint action $\boldsymbol{a}$. 
Our work is in line with ILs, which may be more difficult, but resembles real-world decentralized learning and may be more scalable.

\subsection{Challenges of Independent Learners (ILs)}
Even with perfect observability, ILs are non-Markovian due to unpredictable and unobservable teammates' actions, hence the {\it environment non-stationary problem} \cite{bowling2002multiagent}. 
Previous work has highlighted prominent challenges when applying Markovian methods, such as Q-Learning, to ILs: {\it shadowed equilibria} \cite{fulda2007predicting}, {\it stochasticity}, and {\it alter-exploration} \cite{matignon2012independent}.

{\it Shadowed equilibria} is the main issue we are addressing, which must be balanced with the {\it stochasticity} problem. Without communication, independent learners who are maximizing their expected returns are known to be susceptible to sub-optimal Nash equilibria where the suboptimal joint policy can only be improved by changing all agents' policies simultaneously. To battle this issue methods typically put more focus on high reward episodes, with the hope that all agents will be able to pursue the maximum reward possible, forgoing the objective of maximizing the expected return.

Optimistic methods are more robust to {\it shadowed equilibria}, but give up precise estimation of environment stochasticity. Therefore, these methods can mistake a high reward resulting from stochasticity as a successful cooperation \cite{wei2016lenient}. This challenge is called {\it stochasticity}. In environments where high reward exists at low probability, the agents will fail to approach a joint optimal policy.

The {\it alter-exploration} problem arises from unpredictable teammate exploration. In order to estimate state values under stochasticity, ILs have to consider agent exploration. For learners with an $\epsilon$-greedy exploration strategy, the probability of at least 1 out of $n$ agent  exploring at an arbitrary time step is $1 - (1 - \epsilon)^n$. The alter-exploration problem amplifies the issue of {\it shadowed equilibria} \cite{matignon2012independent}.

\section{Related Work}

In a Dec-POMDP, the reward for each agent depends on the joint action chosen by the entire team $\mathcal{I}$; so an agent will likely be punished for an optimal action due to actions from non-optimal teammates.
Teammates' policies are not only unobservable and non-stationary, but are often sub-optimal due to exploration strategies. As a result, vanilla Q-Learning would be forced to estimate the exploratory dynamics which is less than ideal.
We  first  discuss  related  work  for  adapting  independent  learners for multi-agent domains, and then discuss Implicit Quantile Networks, which we will extend.

\subsection{Hysteretic Q-Learning (HQL)}

Hysteretic Q-Learning (HQL) \cite{matignon2007hysteretic} attempts to improve independent learning by injecting overestimation into the value estimation by reducing the learning rate for negative updates. 
Two learning rates $\alpha$ and $\beta$, named the increase rate and the decrease rate, are respectively used for updating overestimated and underestimated TD error $\delta$: 
\begin{equation}
    Q(s,a) \leftarrow
    \begin{dcases}
        Q(s,a) + \beta  \delta  & \text{if } \delta \leq 0 \\
        Q(s,a) + \alpha \delta & \text{otherwise}
    \end{dcases}
\end{equation}
Hysteretic DQN (HDQN) \cite{omidshafiei2017deep} applies hysteresis to DQN, whose TD error is given by 
\begin{equation}
\delta_t := Q^{\theta_i}(s_t,a_t) - (r + \gamma f\max\limits_{a^{\prime}}Q^{\theta_i^{-}}(s_{t+1},a^{\prime})).
\end{equation}
In practice, HDQN fixes the increase rate $\alpha$ (e.g. $\alpha=0.001$), and scales the decrease rate as $\beta\alpha$.
We thus only discuss the effect of tuning $\beta$. %
In order to reason under partial observability, Hysteretic Deep {\it Recurrent} Q-Networks (HDRQN) \cite{omidshafiei2017deep}, uses a recurrent layer (LSTM) and is trained using replay buffers featuring synchronized agent samples (called CERTs)  \cite{omidshafiei2017deep}. Our work utilizes the same buffer structure.

\subsection{Lenient Deep Q-Network (LDQN)}

Lenient Deep Q-Network (LDQN) \cite{palmer2018lenient} incorporates lenient learning \cite{panait2006lenient} with DQN by encoding the high-dimensional state space into lower dimensions where temperature values are feasible to be stored and updated. 
Leniency, used to determine the probability of negative value updates, is obtained from exponentially decaying temperature values for each {\it state encoding and action} pair using a decay schedule with a step limit $n$, the schedule $\beta$ is given by:
\begin{equation*}
    \beta_t = e^{\rho \times d^t}
\end{equation*}
for each $t$, $0 \leq t < n $, where $\rho$ is a decay exponent which is decayed using a decay rate $d$. The decay schedule aims to prevent the temperature from premature cooling.
Given the schedule, the temperature $T$ is folded and updated as follows:
\begin{equation*}
\begin{split}
    T_{t+1}(\phi(s_t),a_t)=\beta_t\Big((1-\upsilon) T_t(\phi(s_t),a_t) +
    \upsilon \underset{a \in A}{\mathbb{E}}T_t(\phi(s_{t+1}),a)\Big)
\end{split}
\end{equation*}
where $\upsilon$ is a fold-in constant. Then, the leniency of a state-action pair is calculated by look up in the temperature table and given by:
\begin{equation}
    l(s,a) = 1 - e^{-K\times T(\phi(s),a)}
\end{equation}
where $K$ is a leniency moderation constant. %
LDQN schedules optimism injected in state-action estimates, mitigating {\it shadowed equilibria}, and is able to be robust against {\it over optimism} as leniency decreases over time.
On the other hand, successfully applying LDQN requires careful consideration for decay and moderation parameters, whereas our approach requires fewer hyper-parameters and is robust to different parameter values, yet yields higher performance in terms of improved sample efficiency.

\subsection{Implicit Quantile Network (IQN)}
\label{sec:IQN}
IQN \cite{dabney2018implicit} is a single-agent Deep RL method which we extend to multi-agent partially observable settings. As a distributional RL method, quantile networks represent a distribution over returns, denoted $Z^{\pi}$ for some policy $\pi$, where $\mathbb{E}(Z^{\pi})=Q^{\pi}$, by estimating the inverse c.d.f. of $Z^{\pi}$, denoted $F^{-1}_{\pi}$. Implicit Quantile Networks estimate $F^{-1}_{\pi,\tau}(s,a)$ for a given state-action pair, $s,a$, from samples drawn from some base distribution ranging from 0 to 1: $\tau \sim U([0,1])$, where $\tau$ is the quantile value that the network aims to estimate. The estimated expected return can be obtained by averaging over multiple quantile estimates:
\begin{equation}
    Q_{\omega}(s,a) := \underset{\tau \sim U([0,1])}{\mathbb{E}}[F^{-1}_{\pi,\omega(\tau)}(s,a)]
\end{equation}
where $\omega: [0,1]\xrightarrow{}[0,1]$ distorts risk sensitivity. Risk neutrality is achieved when $\omega = \mathds{1}$. In Section \ref{dynamic_risk} we will discuss how we distort risk in multi-agent domains and do so in a dynamic fashion where risk approaches neutral as exploration probability approaches $0$.

The quantile regression loss \cite{koenker2001quantile} for estimating quantile at $\tau$ and error $\delta$ is defined using Huber loss $\mathcal{H}_\kappa$ with threshold $\kappa$
\begin{equation}
\rho_{\tau}(\delta) = (\tau - \mathds{1}_{\delta \leq 0}) \frac{\mathcal{H}_\kappa(\delta)}{\kappa}
\end{equation}
which weighs overestimation by $1-\tau$ and underestimation by $\tau$, $\kappa = 1$ is used for linear loss.
Given two sampled $\tau, \tau^{\prime} \sim \omega(U([0,1]))$ and policy $\pi_{\omega}$, the sampled TD error for time step $t$ follows distributional Bellman operator:
\begin{equation*}
\delta^{\tau,\tau^{\prime}}_t = F^{-1}_{\tau}(s_t,a_t) - (r_t + \gamma F^{-1}_{\tau^{\prime}}(s_{t+1},\pi_{\beta}(s_{t+1}))).
\end{equation*}

Thus, with sampled quantiles $\tau_{1:N}$ and $\tau_{1:N^{\prime}}$, the loss is given by:
\begin{equation}
    L = \frac{1}{N^{\prime}}\sum^{N}_{i=1}\sum^{N^{\prime}}_{j=1}\rho_{\tau_i}(\delta^{\tau_i,\tau^{\prime}_j})
\end{equation}

Distributional learning have long been considered a promising approach due to reduced {\it chattering} \cite{gordon1995stable,kakade2002approximately}. Furthermore, distributional RL methods have shown, in single agent settings, robustness to hyperparameter variation and to have superior sample efficiency and performance \cite{barth2018distributed}.

\section{Our Approach}

We use IQN as the basis of our method since it has shown state-of-the-art performance in single-agent benchmarks, but more importantly, because we believe that learning a distribution over returns provides a richer representation of transitional stochasticity {\it and} exploratory teammates in MARL.
Consequently, the distributional information can be utilized to encourage coordination, but also properly distribute blames among agents, which has historically been difficult to balance.
We propose Time Difference Likelihood (TDL) in this section and Dynamic Risk Distortion (DRD), both utilize distributional information to foster cooperation.

Time Difference Likelihood (TDL) is a granular approach for controlling the learning rate in a state-action specific fashion, but without an explicit encoder. Instead,  TDL measures the likelihood of a return distribution produced by the target network given the distribution produced by the main network. The motivation is twofold: first, for similar distribution estimations, even with drastic difference in specific quantile location, the learning rate should remain relatively high to capture local differences and improve sample efficiency; second, for teammate explorations, TDL will more likely to be low, hence applying more hysteresis on non-Markovian dynamics. Also, as we show from empirical evaluations, TDL acts as a state-specific scheduler which causes the learning rate to increase over time for states which have received enough training, resulting in more recognition of environment stochasticity, thus converging more robustly towards a joint optimal policy.

Dynamic Risk Distortion (DRD), on the other hand, does not impose value overestimation like hysteresis and leniency; instead, DRD controls the way in which policies are derived from value estimates, by distorting the base distribution from which quantile estimation points $\tau$ are sampled. 
Empirically, DRD is robust to different scheduling hyper-parameters, and allows for faster learning and better performances.
Both approaches can be combined to gain better performances.

\subsection{Time Difference Likelihood (TDL)}
\label{sec:tdl}
We first discuss TDL, a measure which we propose to later guide the magnitude of the network's learning rate contingent on each update.
Motivated to reduce the learning rate when encountering exploratory teammates, but properly updating for local mistakes, we would like to find an indicator value distinguishing the two scenarios.
TDL is such an indicator. %
We scale the learning rate using TDL as discussed later in section \ref{tdl-update}.

At a high level, to calculate the TDL, we first sample from estimated return distributions (using both the main network and the target network) for given observation-action pairs. For simplicity, we denote these as $d_{1:M}:=F^{-1}_{\tau_{1:M}}(s_t,a_t)$ and $t_{1:M^{\prime}} := r_t + \gamma \underset{a^{\prime}}{max} F^{-1}_{\tau^{\prime}_{1:M^{\prime}}}(s_t,a^{\prime})$, where $M$ and $M^{\prime}$ are the number of samples drawn from the base distribution. We call them distribution samples and target samples. Observe that obtaining these samples does not add computational complexity, since we can reuse the samples that were used for calculating losses. 

Next, we formalize an approximation method for estimating the likelihood of a set of samples, given a distribution constituted by another set of samples. TDL, in particular, is the likelihood of target samples given the distribution constituted by distribution samples.
We denote the probability density function given by the distribution samples as $\mathcal{P}(X) := P(X \given d_{1:M})$. The intuition of calculating TDL is to treat the discrete distribution samples as a continuous p.d.f. on which the proximity intervals of target samples are calculated for their likelihoods. 
More specifically, if given $\mathcal{P}$, we estimate the likelihood of target samples as follows:
\begin{equation}
    l_{t_{1:M^{\prime}},d_{1:M}} = \underset{j \in 1:M^{\prime}}{\sum}\mathcal{P}\Big(\frac{t_{j-1}+t_j}{2} \leq X \leq \frac{t_j+t_{j+1}}{2}\Big).
\end{equation}
Now we only need an approximation of the continuous p.d.f. $\mathcal{P}$ which is represented by discrete samples. Our continuous representation is constructed by assuming the density between neighboring samples $d_i$ and $d_{i+1}$ is linear for generalizability and implementation simplicity.
We therefore obtain a set of continuous functions $F_i(X)$ each with domain $(d_i,d_{i+1}]$, where $F_i$ denotes a linearity fits $(d_i, \tau_i)$ and $(d_{i+1}, \tau_{i+1})$.

Let $\mathcal{F}(X) = F_i(X)$ iff $X\in (d_i,d_{i+1}]$. In other words, $\mathcal{F}$ is obtained by connecting all the distribution samples into a continuous monotonically increasing probability density function, which consists of $M-1$ connected linear segments.
Using $\mathcal{F}$ as the c.d.f approximation for $\mathcal{P}$, by definition, for arbitrary $a$ and $b$: $g $,
which can be obtained using the linearity property we defined for $\mathcal{F}$:
\begin{equation}
 \mathcal{P}(a < X \leq b)   = \sum_{i=1}^{M-1}\frac{|(a,b]\cap(d_i,d_{i+1}]|}{d_{i+1}-d_i}(\tau_{i+1} - \tau_i).
\end{equation}

Note that intervals $(-\infty,d_1] $ and $(d_i,\infty]$ have no probability density, hence are omitted.
TDL can be calculated using an arbitrary number of samples for all $M>1$ and $M^{\prime}>0$. 

We can view TDL as not only a noisy consistency measurement between the main and target networks, but also an indicator of information sufficiency in the return distribution estimation. The latter is important for training MADRL agents because it aims to differentiate stochasticity from  non-stationary, which allows agent to obtain a more faithful value estimates based on the environment alone, not other peers.

\subsection{Likelihood Hysteretic IQN (LH-IQN)}
\label{tdl-update}
Hysteretic Learning \cite{matignon2007hysteretic} incorporates low returns in a delayed fashion, by updating value estimations at a slower rate when decreasing.
Hysteretic approaches show strong performance in both tabular and deep learning evaluations, yet fail to delay value estimations synchronously across the  state-action space.
Leniency \cite{panait2006lenient} addresses this issue by recording temperature values in the state-action space. 
Temperature values control the negative update probability, which decrease when an update happens to the corresponding state-action pair. However, when applied in large or continuous state and action spaces, not only is state-action encoding required for computational tractability, but extra care is required for scheduling the temperature \cite{palmer2018lenient}; Palmer et al.~found it necessary to apply temperature folding techniques to prevent the temperature from prematurely extinguishing.

To combat these issues, we introduce Likelihood Hysteretic IQN (LH-IQN) which incorporates TDL with hysteretic learning.
Intuitively, LH-IQN is able to automatically schedule the amount of leniency applied in the state-action space without careful tuning of temperature values thanks to state-action specific TDL measurements.
While deep hysteretic learning uses $0 < \beta <\alpha \leq 1$ to scale learning rates, our LH-IQN uses the $max$ of $\beta$ and TDL as the {\it decrease rate}. More specifically, the learning rate $\mu_t$ is given by:
\begin{equation}
\label{eq:update}
    \mu_t =
    \begin{dcases}
        max(\beta, l_{t_{1:M^{\prime}},d_{1:M}}) \bar \mu, & \text{if } \delta^{\tau,\tau^{\prime}}_t \leq 0\\
        \bar \mu,    & \text{otherwise}
    \end{dcases}.
\end{equation}
where $\bar \mu$ is a base learning rate suitable for learning assuming a stationary environment (e.g. 0.001), $l_{t_{1:M^{\prime}},d_{1:M}}$ is the likelihood defined in Section \ref{sec:tdl} and  $\delta^{\tau,\tau^{\prime}}_t$ is the TD error defined in Section \ref{sec:IQN}.
To explore the effect of likelihood and hysteresis during evaluation, we also define L-IQN as an IQN architecture which only uses TDL $l_{t_{1:M^{\prime}},d_{1:M}}$ as the decrease rate, and H-IQN which only uses $\beta$ as the decrease rate.
Empirically, $\beta$ ranging from $0.2$ to $0.4$ yields high performance.

Since TDL generally increases as the network trains toward consistency, the amount of optimism/overestimation added by hysteretic updates is reduced over time, which is analogous to leniency. 
The key difference is that for domain non-stationarity (caused by stochasticity and/or shifts in teammate policies), which remains unpredictable forever, TDL remains small, effectively employing a low learning rate toward such transitions.

\subsection{Dynamic Risk Sensitive IQN}
\label{dynamic_risk}
Distributional RL has also been studied for designing risk sensitive algorithms \cite{morimura2010nonparametric}. 
We introduce dynamic risk sensitive IQN which utilizes what we call {\it dynamic risk distortion operators}.
IQN has shown to be able to easily produce risk-averse and risk-seeking policies by integrating different {\it risk distortion measures} $\omega: [0,1] \rightarrow [0,1]$ \cite{yaari1987dual, dabney2018implicit}.
In single agent positive-sum games, risk-averse policies are sometimes preferred to actively avoid terminal states for more efficient exploration. 
In MARL, however, agents may benefit from risk-seeking policies as seeking the highest possible utility helps the team break out of sub-optimal shadowed equilibria. 
As we are not boosting the value estimations directly, we say this approach injects {\it hope} instead of optimism.
In our work, we let IQN learn to reflect the true perceived domain dynamics (no learning rate adjustments), but consider generally higher quantile locations (larger values) when making decisions, producing optimistic policies without raising value estimations. 
Again, to be robust to environment stochasticity, we anneal the amount of distortion we apply so that in the end we produce policies based on realistic (non-optimistic) value estimations. We discuss two such distortion operators: CVnaR and Wang \cite{wang2000class}.

CVnaR, Conditional Value-not-at-Risk, is inspired by well studied risk-averse operator Conditional Value-at-Risk ($\text{CVaR}(\eta,\tau) = \eta\tau$) \cite{chow2014algorithms}. Our CVnaR is defined as follows:
\begin{equation}
\text{CVnaR}(\eta, \tau) = 1 - \eta\tau.
\end{equation}CVnaR maps $\tau \sim U([0,1])$ to $\text{CVnaR}(\eta,\tau) \sim U([\eta,1])$, and as $\eta$ reduces, CVnaR become less risk-seeking.

Wang \cite{wang2000class} is a distortion operator whose range always remains $[0,1]$, but becomes exponentially increasing (probability density shifted towards $1$) when given positive bias parameter $\eta$. Wang is defined as: 
\begin{equation}
\text{Wang}(\eta,\tau) = \Phi(\Phi^{-1}(\tau) + \eta)
\end{equation}
where $\Phi$ is the standard Normal cumulative distribution function. Observe that when $\eta \rightarrow 1$, Wang almost always returns $1$, becoming the most risk-seeking distortion operator possible. Also, like CVnaR, as $\eta\rightarrow 0$, risk-neutrality is observed.

We found it suitable to linearly anneal $\eta$ (for both Wang and CVnaR) during training to achieve better stability as the agent becomes more and more risk-neutral, but behaves like a maximization approach in the beginning. The aim is that during the initial risk-seeking period when $\eta$ is high, agents are encouraged to explore highly rewarding spaces, which supports them to better break out of shadowed equilibra; whereas in the end, the risk-neutral distortion produces an unbiased policy which is unlikely to fall for domain stochasticity. We show that, empirically, risk distortion not only improves overall performance when applied alone, but also when used in conjunction with TDL.

\section{Evaluation}

In this section, we compare our methods with previous state-of-the-art methods in multiple domains. We begin by comparing likelihood hysteretic IQN (LH-IQN) with the previous state-of-the-art, HDRQN and LDQN, and then analyze the effect of TDL as well as risk distortion operators.
Results shown in all Figures use decentralized training with 20 random seeds.
\subsection{Evaluation on meeting-in-a-grid}

We first conduct experiments on a standard partially observable meeting-in-a-grid domain \cite{amato2009incremental} to be consistent with previous work \cite{omidshafiei2017deep}.
The meeting-in-a-grid task consists of one moving target and two agents in a grid world. Agents get reward of 1 for simultaneously landing on the target location and 0 otherwise. Episodes terminate after 40 transitions or upon successful meeting at the target. Agents have noisy transition probability of 0.1, where agents' moves end up in an unintended position (uniformly left, right and still) at the rate of $0.1$. Observations include flickering locations of the agents themselves and the target. 

Again, to be consistent with previous work~\cite{omidshafiei2017deep}, we use recurrent versions of our methods.   
We label the architecture with added Recurrency as LH-I{\bf R}QN.
The network starts with 2 fully connected layers of 32 and 64 neurons respectively, then has an LSTM layer with 64 memory cells and a fully connected layer with 32 neurons which then maps onto value estimates for each action.
We use $\beta=0.4$, $\gamma = 0.95$ and Adam \cite{kingma2014adam} for training. 
For quantile estimators, we sample 16 for $\tau$ and $\tau^{\prime}$ to approximate return distributions, and $\tau$ embeddings are combined with the LSTM output. 

\begin{figure*}
\hspace{-10pt}
  \subfigure[Performance on $4\times4$ meeting-in-a-grid benchmark]{\includegraphics[width=.5\textwidth]{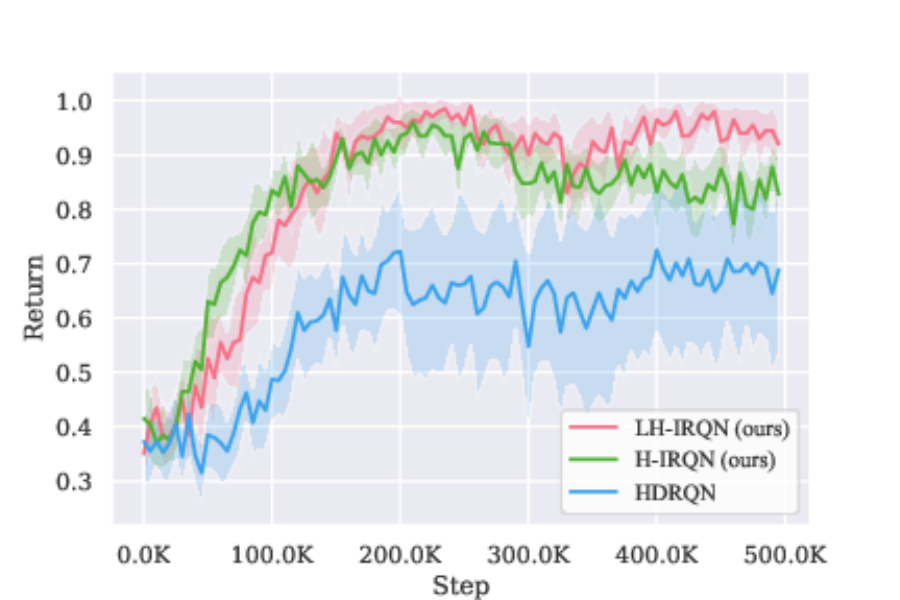} \label{fig:capture_flag}}
  \subfigure[TDL Value]{\includegraphics[width=.5\textwidth]{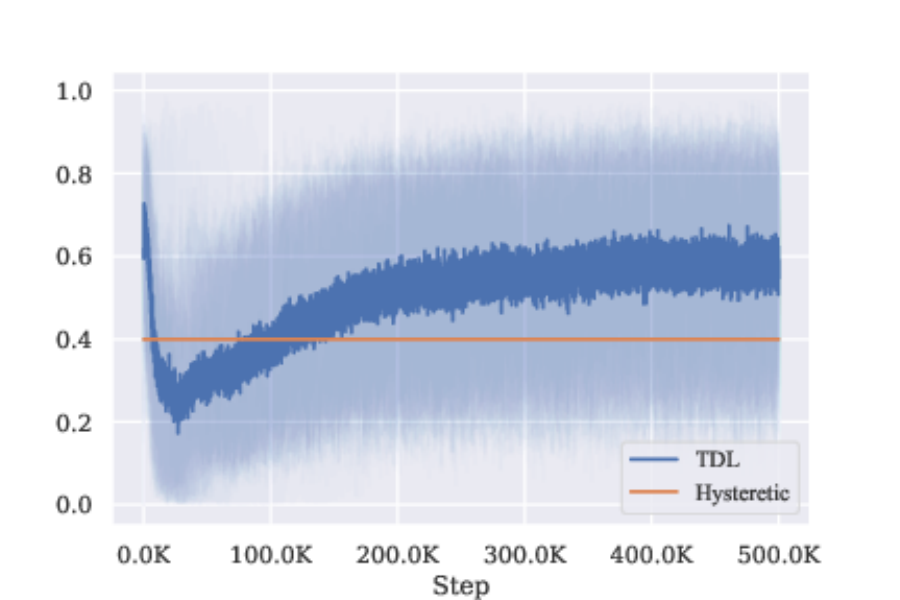}\label{fig:hys_value}}
  \caption{(a) IRQN models perform better than HDRQN, especially with TDL (b) TDL values during training of LH-IRQN, shows clear increase of usage of TDL over hysteresis.}
\end{figure*}

We first evaluate LH-IRQN's performance against HDRQN \cite{omidshafiei2017deep} and H-IRQN on a 4$\times$4 grid (Fig.~\ref{fig:capture_flag}). H-IRQN is a version of LH-IRQN that does not use TDL, but uses IRQN with hysteresis (Eq~\ref{eq:update}).
Both of our IRQN methods outperform HDRQN in both learning rate and final performance. 
HDRQN has a large variance because it does not robustly solve the task---only a portion of seeds reached near-optimal policies.
Our IRQN-based methods show more stability concerning reaching optimality, but not utilizing TDL makes agents susceptible to environment stochasticity, producing fewer near-optimal joint policies over time. 
Our methods similarly outperform HDRQN in higher dimensional ($5\times5$, $6\times6$) variations of the benchmark 
, except for $3\times3$ which is too simple to differentiate the methods.

Directly applying LDQN, with convolution layers replaced by fully-connected layers to better suit the observations, on meeting-in-a-grid failed to solve the tasks due to the high flickering probability and the observation encoding. 
Additional comparisons with LDQN are given in 
section \ref{sec:high_d} and  \ref{sec:cmotp}, but this shows the sensitivity of LDQN to the task and encoding.

As shown in Fig.~\ref{fig:hys_value}, TDL increases over time during training, while maintaining a high variance which resulted from domain non-stationary as expected. 
Overall, the usage of TDL versus hysteresis $\beta$ increases significantly; as TDL is used when it is larger than $\beta$, which can be considered a lower cap. The overall learning rate for negative samples (i.e., those with non-positive TD error) is increased over time, thus adding less hysteresis and optimism to experiences deemed predictable by TDL, eventually theoretically learning unbiased state value estimations  \cite{matignon2007hysteretic}.
While one would expect methods with less optimism to be susceptible to action shadowing, our Likelihood method nonetheless achieves better stability and performance as shown in Fig. \ref{fig:capture_flag}, from which we can deduce that TDL is able to distinguish domain non-stationary from stochasticity as we theorized.
The spike (and dip) at the beginning seen in Fig~\ref{fig:hys_value} is due to immature quantile estimations being used to calculate TDL; during the start of training, these quantile values are not guaranteed to represent a valid distribution---they may be aggregated together or even reversed depending on the network weight initializations.
As a result, it is unstable to solely use TDL as a decrease rate, a problem which we solved with maximizing with hysteresis parameter $\beta$, which is essentially a lower bound to prevent the network from  terminating the learning process when estimations differ drastically from target estimates. This issue can also also be  mitigated using Dynamic Risk Distortion (DRD) which can be used to achieve extremely optimistic distortion during the beginning phase of training. We discuss empirical improvements of DRD in Section \ref{dynamic_risk}.

\subsection{Multi-Agent Object Transportation Problems (CMOTPs)}
\label{sec:cmotp}

We also evaluate LH-IQN on variations of Coordinated Multi-Agent Object Transportation Problems (CMOTPs) \cite{palmer2018lenient}, consistent with Palmer et al.'s work on LDQN. CMOTPs require two agents carrying a box to a desired location for a terminal reward; the box moves when agents are adjacent and move in the same direction. Variations of the task include obstacles and stochastic rewards. CMOTPs have $16 \times 16$ observations with added noise. 

Our network architecture mimics that of LDQN for comparability: two convolutional layers with 32 and 64 kernels, a fully connected layers with size 1024 which combines quantile embedding, followed by another fully connected layer with size 1024, which then maps onto value estimates for each action.
Hyper-parameters remain the same as  original work which were found suitable for training in CMOTPs.

\begin{figure}
\includegraphics[width=.5\textwidth]{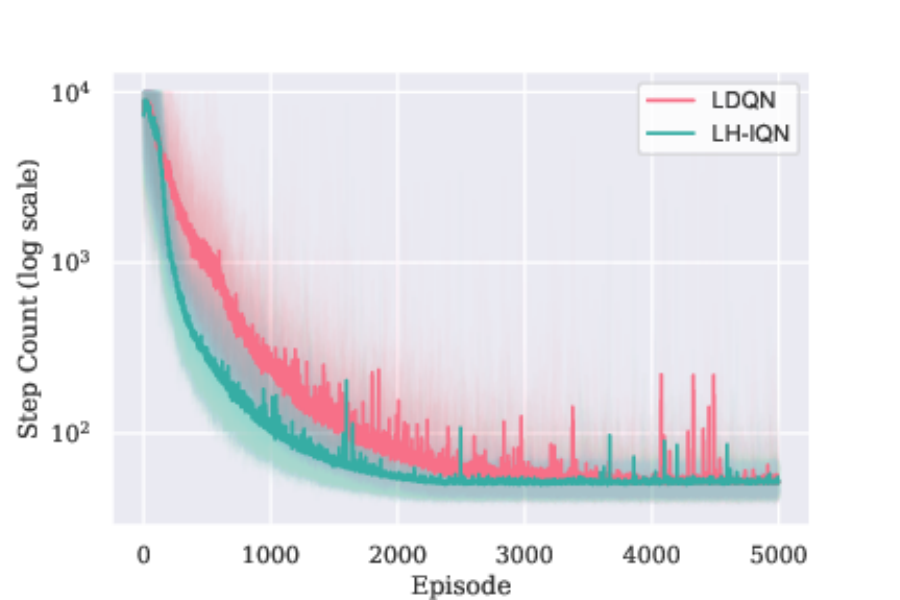}
  \caption{CMOTP benchmark results with a total of 60 runs, aggregated over all three CMOTP variants.}
  \label{fig:cmotp}
\end{figure}

\begin{figure*}
\hspace{-10pt}
  \subfigure[4-agent high-dimensional meeting-in-a-grid benchmark]{
    \includegraphics[width=.5\textwidth]{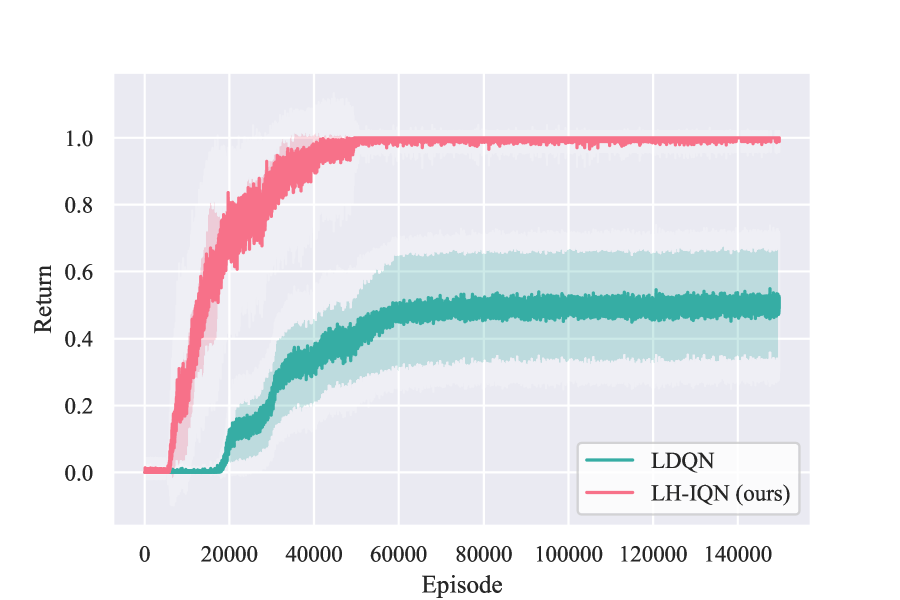} \label{fig:4_agents_rewards}}
\hspace{-10pt}
  \subfigure[5-agent high-dimensional meeting-in-a-grid benchmark]{
    \includegraphics[width=.5\textwidth]{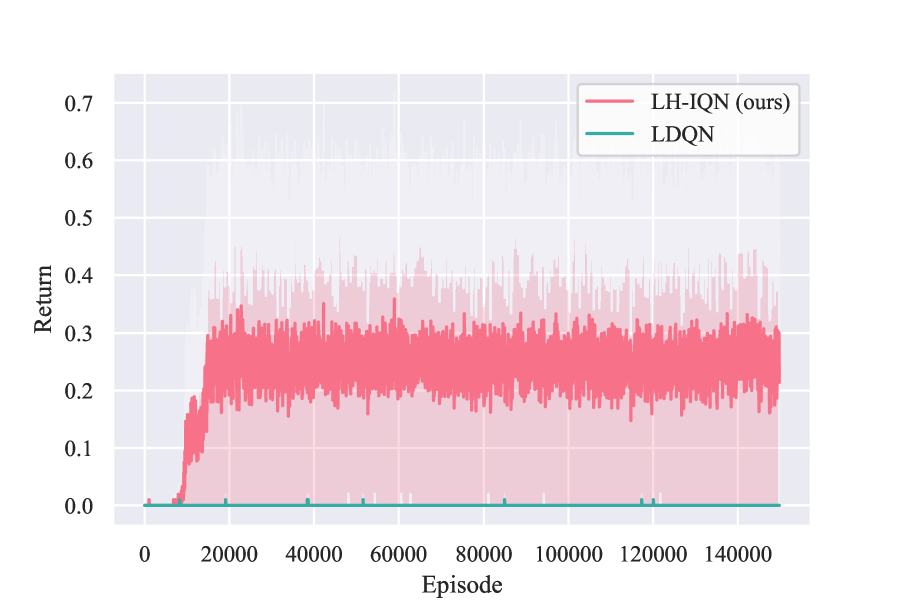}\label{fig:5_agents_rewards}}
  \caption{High dimensional meeting-in-a-grid 4×4 benchmark with more agents.}
   \vspace{.1in}
\end{figure*}

As seen in Fig. \ref{fig:cmotp}, although both methods converge to a policy that solves the problem consistently, our method shows an improved sample efficiency.
We hypothesize that the temperature is decaying less aggressively than it should be in LDQN, which is likely due to temperature folding techniques and/or that the hashing space of the autoencoder is larger than the theoretical minimum.

On the other hand, our method utilizes TDL to scale negative updates and shows better sample efficiency.
Initially the value estimations do not seem optimistic enough to perform coordinated actions or to propagate to an earlier-stage state, but the likelihood estimation has the added benefit of being able to produce small values in under-explored state-action space, while hesitating less to update negatively in explored spaces.
TDL also helps to synchronize optimism across state-action space; in other words, the ability to estimate a distribution consistency adds less optimism to state-action pairs which have received enough training to be able to produce consistent distributions.

\begin{figure}
\includegraphics[width=.45\textwidth]{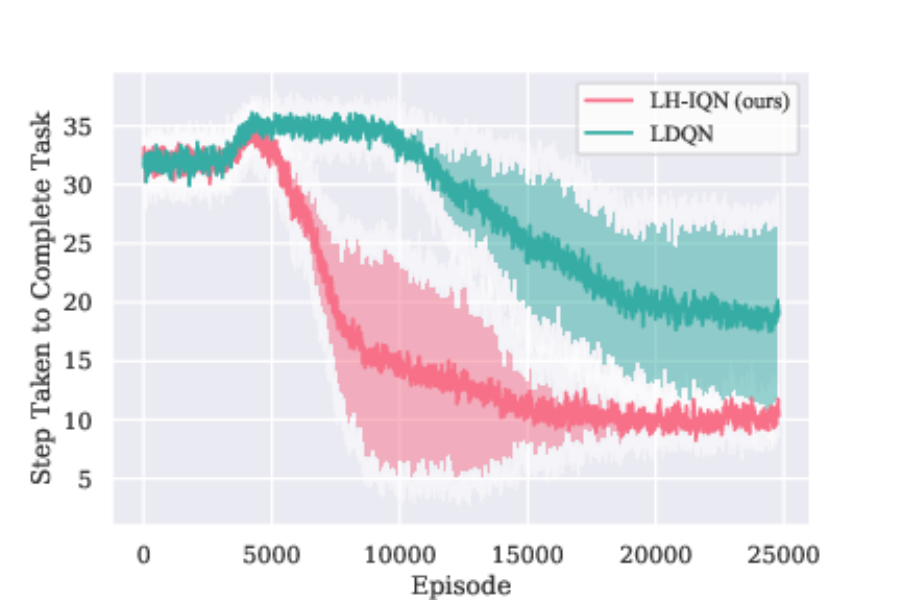}
  \caption{Evaluation of LH-IQN compared with LDQN on the high-dimensional meeting-in-a-grid 4$\times$4 benchmark. Figure shows the number of steps taken to complete tasks (small values preferred).}
  \label{fig:hi_d_capture}
 
\end{figure}

\subsection{High dimensional meeting-in-a-grid task}
\label{sec:high_d}

Motivated to most fairly compare the performance of our approach with LDQN and to compare on more than two agents, we modify the existing $4\times4$ meeting-in-a-grid benchmark to produce graphical observations ($16\times16$) with added noise, a type of task on which LDQN is originally evaluated.
Due to the difficulty of grid searching numerous hyper-parameters for LDQN, the parameters used were linearly searched individually while fixing others based on the original work.
We found reducing temperature schedule decay rate $d$ from $0.9$ to $0.8$ helps with convergence in our task, possibly due to meeting-in-a-grid's shorter scenarios.
We used: $K=3.0, d = 0.8, \xi=0.25$ and $\mu=0.9995$ along with the autoencoder, where $\xi$ is the exponent for temperature-based exploration, and $\mu$ is the decrease rate for maximum temperature.

As seen in Fig.~\ref{fig:hi_d_capture}, our method shows higher sample efficiency and performance. Noticing the y-axis is the number of steps needed to complete the task. We see that LDQN was able to solve the task, however it is not as stable and has less ideal performance compared to LH-IQN. 
As the task becomes reliably solvable, LDQN slows down learning and has a high variance, whereas LH-IQN achieves the optimal solution on every run. 
We notice that the temperature values of LDQN are low during the final stages of training, suggesting minimal leniency is applied. Therefore, it appears the joint policy reaches a shadowed equilibrum with less effective explorations.

\begin{figure}
    \includegraphics[width=0.45\textwidth]{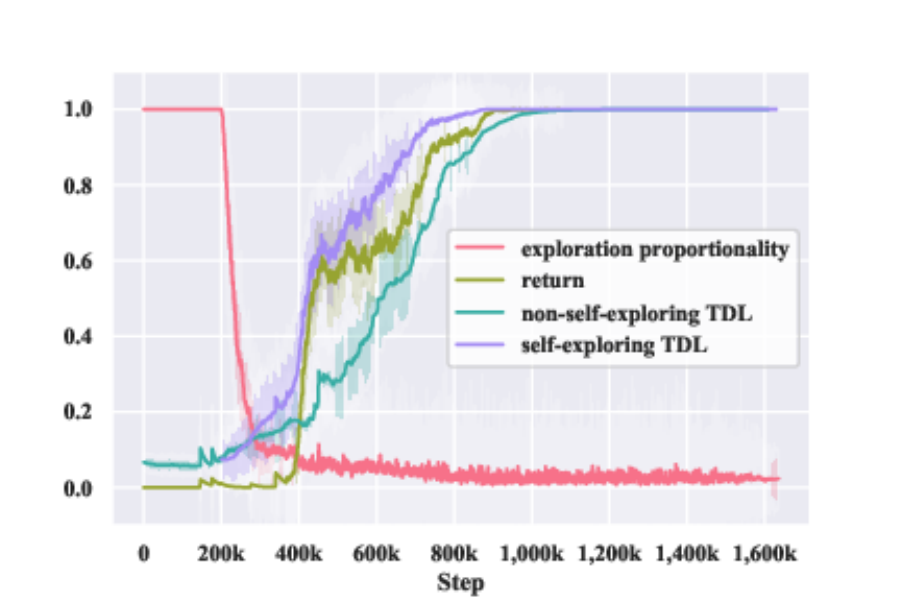}
    \caption{TDL values in different exploration situations, where self-exploring TDL refers to TDLs produced from training sample transitions in which the agent is exploring, and non-self-exploring TDL refer to teammates' explorations. Exploration proportionality shows how much agents are actively exploring.}
    \label{fig:tdl_analysis}
\end{figure}

\paragraph{Benchmarking with more than two agents.}
Since the probability of effective explorations decreases exponentially as the number of agents increases due to {\it alter-exploration} \cite{matignon2012independent}, independent learners often suffer from poor scalability in the number of agents. LH-IQN mitigates this issue by putting more emphasis on non-exploratory episodic samples which we show in ~\ref{tdl_trend} and Fig.~\ref{fig:tdl_analysis}.
We evaluate our method against LDQN in those scenarios involving more agents; and as shown in Fig.~\ref{fig:4_agents_rewards}, LDQN failed to solve the 4-agent environment reliably, persisting at a $50\%$ fail rate, whereas our method's fail rate converges to 0 (return of 1).
Moreover, in the 5-agent environment, shown in Fig.~\ref{fig:5_agents_rewards} LDQN failed to learn any effective policies even though it does encounter cooperatively successful episodes. On the contrary, while noisy, our method is able to successfully learn in this large domain. 

\paragraph{Explorations and TDL values}
\label{tdl_trend}
We inspect the TDL values during training the 4 agent high-dimensional meeting-in-a-grid task and plot training TDL values in Fig.~\ref{fig:4_agents_rewards} depending on whether the agent was exploring or not.
The TDL trends are shown in Fig.~\ref{fig:tdl_analysis}.
{\it Self-exploring TDL} refers to TDL values produced from training batch transitions in which the agent is actually exploring (uniformly action selection as opposed to greedily maximizing expected return), and {\it non-self-exploring TDL} refers to that of when teammates were exploring but the agent is not.
We observe that self-exploring TDL values are much higher than that of non-self-exploring during the active policy improvement period.
This divergence of TDL values when the agent is versus is not exploring suggests that TDL is able to distinguish local mistakes (where it applies higher learning rate) from teammate explorations as we theorized.
Note that the agent does not have access to ground-truth exploration information in any stage of training, therefore TDL is inferring exploratory information only from the given transitions.

\subsection{Dynamic Risk Sensitive IQN}

\begin{figure}
    \includegraphics[width=0.45\textwidth]{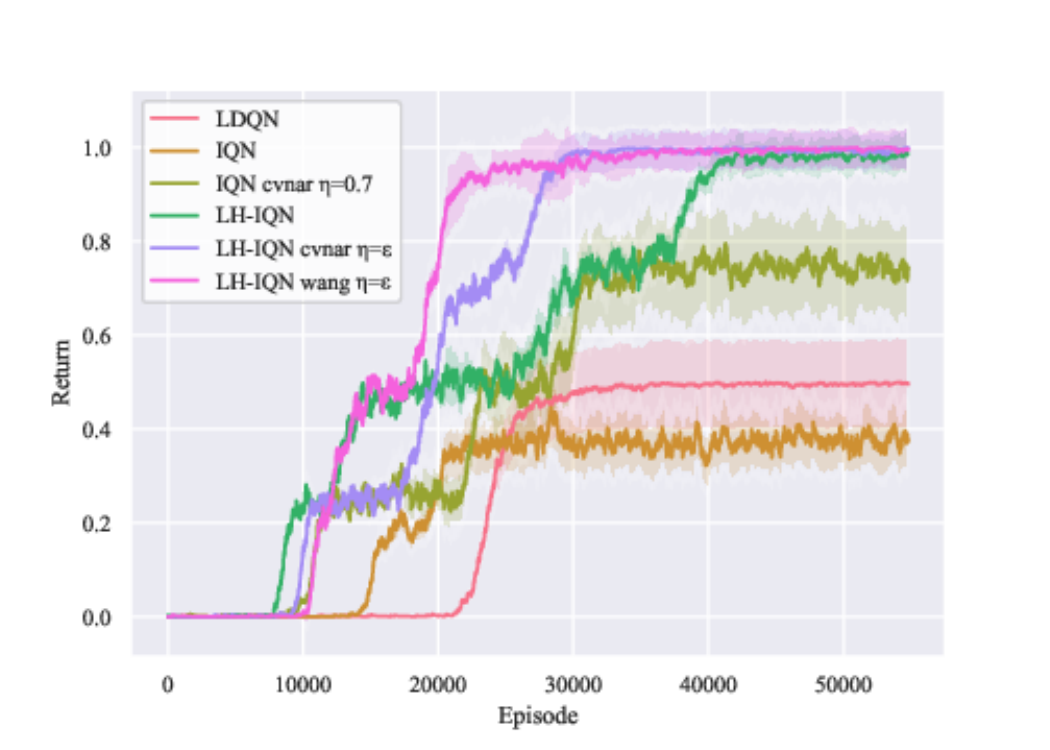}
    \vspace{-.1in}
    \caption{Performance of Risk Distortion methods applied to IQN with and without Likelihood-Hysteresis on the $4\times4$ high-dimensional meeting-in-a-grid benchmark with high environment stochasticity (sliding probability of $0.3$ for moving); shows that distortion operator works well with and without TDL.}
    \label{fig:risk_distortion}
\end{figure}

We also demonstrate usage of dynamic risk distortion (DRD) operators to produce optimistic policies in an environment with even more environment stochasticity, highlighting that DRD can be used in conjunction with TDL for policy stability.
As shown in Fig.~\ref{fig:risk_distortion}, both operators, CVnaR and Wang with $\eta=\epsilon$, when applied on top of our IH-IQN approach, reach optimality more efficiently.
Notice that the environment is more challenging as the sliding probability (probability of ending up in an unintended adjacent position upon moving) increased from the previously used 0.1 to 0.3.
TDL is unstable initially depending the weight initializations, often taking on extreme values such as $0$ or $1$ in practice.
Therefore, we reason that the TDL-based agents are likely to fall for environment stochasticity at the beginning of training like previous methods because value estimations across states are not in the same learning stage initially.
However, risk distortion solves the issue that TDL can take on extremely low TDL value in early stages of learning; since in the early stage, high $\eta$ shifts the density of the distribution from $\tau$ towards $1$, making the policy resemble a maximization-based approach, yet still learning in an unbiased manner in terms of value estimations.
We also found that LH-DQN is robust to different $\eta$ values when using both Wang and CVnaR.
We simply used the exploration parameter $\epsilon$ as the value for $\eta$ for our dynamic distortion operator in our evaluation.
Instead of using $\epsilon$, a separate scheduling can be adapted for $\eta$, and we found it to be more appropriate to linearly anneal $\eta$ from $0.9$ to $0.4$, but we found that performance differences are small in the benchmarking environment. 

Also shown in Fig.~\ref{fig:risk_distortion} is that vanilla IQN yields limited performance, but exceeds LDQN when simply applying a fixed  risk distortion (CVnaR) without annealing.
Overall, we observe that DRD usually leads to faster policy improvements due to initially high $\eta$ making it overly optimistic, breaking shadowed equilibria and reducing the initial inconsistencies in quantile estimates; on the other hand, if applied without annealing, the agents' policies are subjected to environment stochasticity, since the derived policies are not maximizing expected return.
Therefore, it would be advisable to reduce (either annealing $\eta$ or reduce the likelihood of distortion) or turn off risk distortion in later stages of training.

\section{Conclusion}

This paper describes a novel distributional RL method for improving performance in cooperative multi-agent reinforcement learning settings. In particular, we propose a likelihood measurement applicable in distributional RL, TDL, that is used for comparing return distributions in order to adaptively update an agent's value estimates.
Through inspecting TDL values and usages, we conclude that TDL plays a part in distinguishing domain non-stationary (e.g., from other agent learning and exploration) and domain stochasticity (including teammate policy shifts), a long standing difficulty.
We compare and analyze our method along side state-of-the-art methods on various benchmarks, and our approach demonstrates improved stability, performance and sample efficiency.
Furthermore, we demonstrate the effectiveness and adaptiveness of our method when incorporating dynamic risk distortion operators, and show risk distortion can also be applied to foster cooperation even without incorporating TDL.

\section{Acknowledgements}
We gratefully acknowledge funding from ONR
grant N00014-17-1-2072 and NSF grant \#1816382. 

\bibliographystyle{ACM-Reference-Format}  %
\bibliography{sample-bibliography}  %
\newpage

\section{Appendix}
\subsection{Results of Individual Variations of CMOTP}

\begin{figure}[h]
  \begin{center}
  \includegraphics[width=\columnwidth]{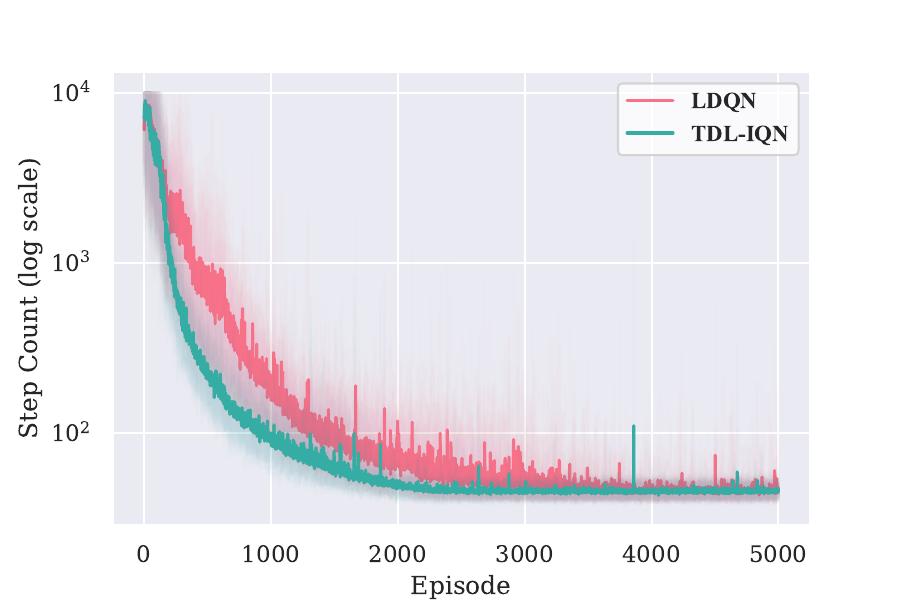}
  \vspace{-20pt}
  \caption{CMOTP Version 1}
  \label{fig:cmotp1}
  \end{center}
\end{figure}

\begin{figure}[h]
  \begin{center}
  \includegraphics[width=\columnwidth]{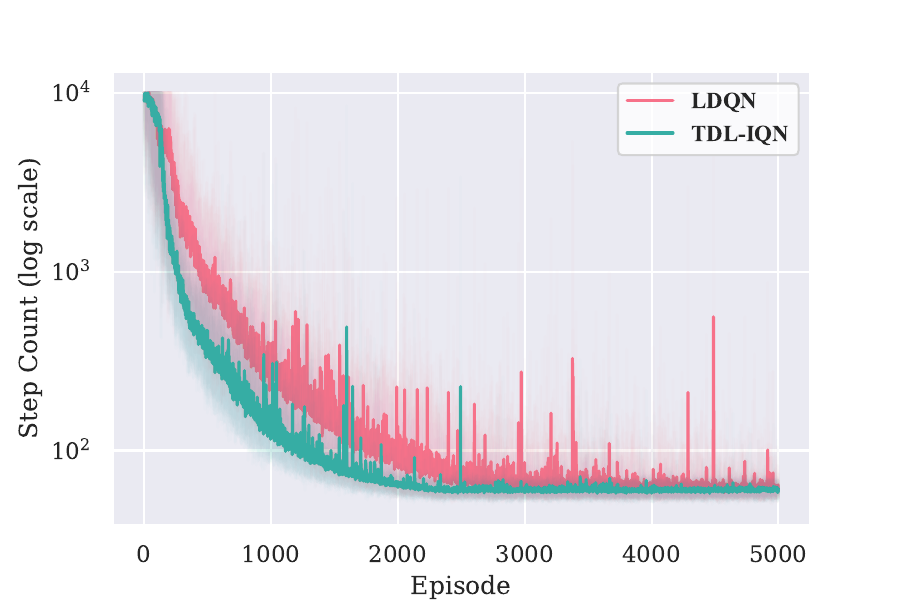}
  \vspace{-20pt}
  \caption{CMOTP Version 2 (Narrow Corridor)}
  \label{fig:cmotp2}
  \end{center}
\end{figure}

\begin{figure}[h]
  \begin{center}
  \includegraphics[width=\columnwidth]{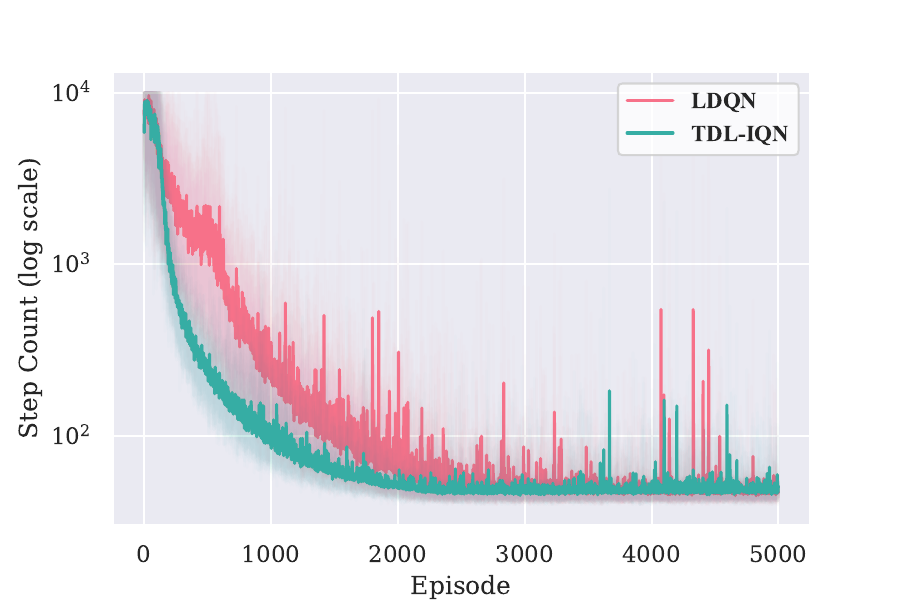}
  \vspace{-20pt}
  \caption{CMOTP Version 3 (Stochastic Reward)}
  \label{fig:cmotp3}
  \end{center}
\end{figure}

\subsection{Results of Variations of Meeting-in-a-Grid}

\begin{figure}[h]
  \begin{center}
  \includegraphics[width=\columnwidth]{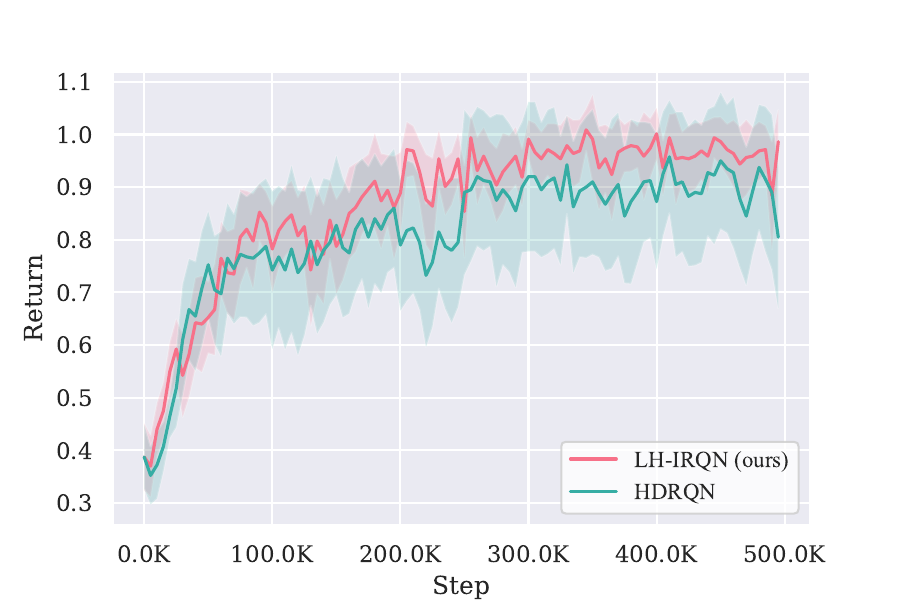}
  \vspace{-20pt}
  \caption{Meeting-in-a-Grid $3\times 3$ benchmark}
  \label{fig:three_by_three}
  \end{center}
\end{figure}

\begin{figure}[h]
  \begin{center}
  \includegraphics[width=\columnwidth]{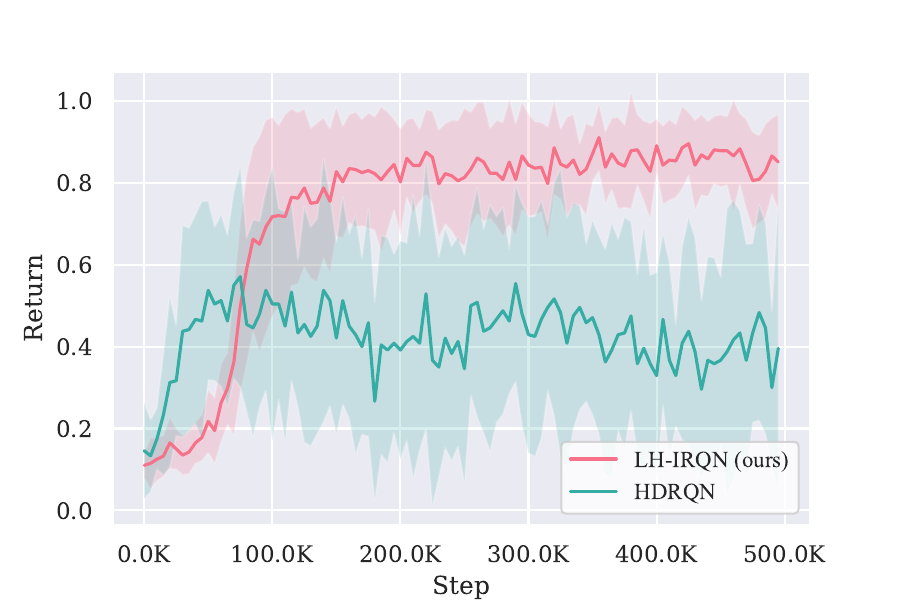}
  \vspace{-20pt}
  \caption{Meeting-in-a-Grid $5\times 5$ benchmark}
  \label{fig:five_by_five}
  \end{center}
\end{figure}

\begin{figure}[h]
  \begin{center}
  \includegraphics[width=\columnwidth]{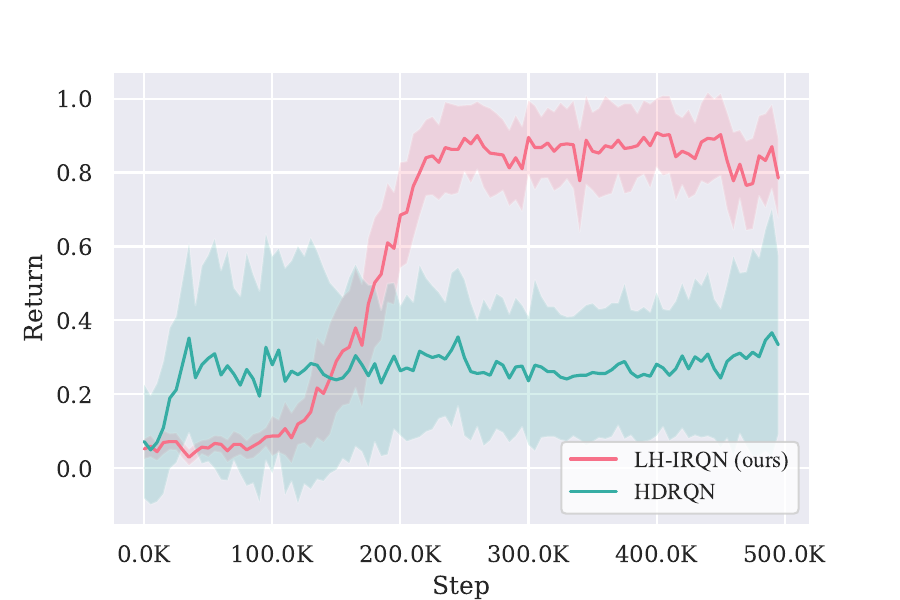}
  \vspace{-20pt}
  \caption{Meeting-in-a-Grid $6\times 6$ benchmark}
  \label{fig:six_by_six}
  \end{center}
\end{figure}

\end{document}